\documentclass[nonacm, sigconf]{acmart}

\AtBeginDocument{%
  \providecommand\BibTeX{{%
    \normalfont B\kern-0.5em{\scshape i\kern-0.25em b}\kern-0.8em\TeX}}}

\setcopyright{none}
\usepackage{tabulary}
\usepackage[normal]{threeparttable}

\begin{document}

\title{The Case for Globalizing Fairness: A Mixed Methods Study on Colonialism, AI, and Health in Africa}
\author{Mercy Asiedu}
\authornote{Both authors contributed equally to this work.}
\affiliation{%
    \institution{Google Research}
    \country{}
}
\email{masiedu@google.com}

\author{Awa Dieng}
\authornotemark[1]
\affiliation{%
    \institution{Google DeepMind}
    \country{}}
\email{awadieng@google.com}

\author{Iskandar Haykel}
\affiliation{%
    \institution{}
    \country{}}
\authornote{Work done while at Google}
\email{alexander.haykel@gmail.com}

\author{Negar Rostamzadeh}
\affiliation{%
    \institution{Google Research}
    \country{}}
\email{nrostamzadeh@google.com}

\author{Stephen Pfohl}
\affiliation{%
    \institution{Google Research}
    \country{}}
\email{spfohl@google.com}

\author{Chirag Nagpal}
\affiliation{%
    \institution{Google Research}
    \country{}}
\email{chiragnagpal@google.com}

\author{Maria Nagawa}
\affiliation{%
    \institution{Duke University}
    \country{}}
\email{maria.nagawa@duke.edu}

\author{Abigail Oppong}
\affiliation{%
    \institution{Ghana NLP}
    \country{}}
\email{abigoppong@gmail.com}

\author{Sanmi Koyejo}
\affiliation{%
    \institution{Google DeepMind}
    \country{}}
\email{sanmik@google.com}

\author{Katherine Heller}
\affiliation{%
    \institution{Google Research}
    \country{}}
\email{kheller@google.com}

\renewcommand{\shortauthors}{Asiedu and Dieng et al.}

\begin{abstract}
With growing application of machine learning (ML) technologies in healthcare, there have been calls for developing techniques to understand and mitigate biases these systems may exhibit. Fairness considerations in the development of ML-based solutions for health have particular implications for Africa, which already faces inequitable power imbalances between the Global North and South. This paper seeks to explore fairness for global health, with Africa as a case study. We conduct a scoping review to propose axes of disparities for fairness consideration in the African context and delineate where they may come into play in different ML-enabled medical modalities. We then conduct qualitative research studies with 672 general population study participants and 28 experts in ML, health, and policy focused on Africa to obtain corroborative evidence on the proposed axes of disparities. Our analysis focuses on {\it colonialism} as the attribute of interest and examines the interplay between artificial intelligence (AI), health, and colonialism. Among the pre-identified attributes, we found that colonial history, country of origin, and national income level were specific axes of disparities that participants believed would cause an AI system to be biased. However, there was also divergence of opinion between experts and general population participants. Whereas experts generally expressed a shared view about the relevance of colonial history for the development and implementation of AI technologies in Africa, the majority of the general population participants surveyed did not think there was a direct link between AI and colonialism. Based on these findings, we provide practical recommendations for developing fairness-aware ML solutions for health in Africa.

\end{abstract}

\keywords{Artificial Intelligence, Machine Learning, Algorithmic Fairness, Colonialism, Health, Africa}

\maketitle

\section{Introduction}
\label{sec:intro}
 
Machine learning (ML) models have the potential for far reaching impact in health. However they also have the potential to propagate biases that reflect real world historical and current inequities and could lead to unintended, harmful outcomes, particularly for vulnerable populations \citep{huang22, siala22, chenIY21, char18, gianfrancesco18, obermeyer19}. 

In recent years, the algorithmic fairness literature \citep{barocas-hardt-narayanan} has proposed different techniques to understand, evaluate, and mitigate algorithmic biases that may cause harm or contribute to unfair or biased decision making throughout the machine learning development and deployment pipeline. However, these have been mostly contextualized within western notions of discrimination.

\paragraph{Fairness in Global health:} Fairness is especially important for global health, which  has been plagued with inequitable power imbalances between high-income countries (HICs) and low- and middle-income countries (LMICs) \citep{holst20}. There are multiple definitions of global health but we use the definition proposed by \cite{beaglehole10}: \textit{"[Global health is defined as] collaborative trans-national research and action for promoting health for all."}
The recent decolonizing global health movement sheds light on ``how knowledge generated from HICs define practices and informs thinking to the detriment of knowledge systems in LMICs''\citep{eichbaum21}. 
Eichbaum et al. \citep{eichbaum21}  explore intersections between colonialism, medicine and global health, and how colonialism continues to impact global health programs and partnerships. 
Given that most machine learning models are developed with problem formulation, resources and data from HICs, and may be imported with little regulation to LMICs, there is a risk for algorithmic colonialism and oppression, detailed extensively by \citep{abeba20} and \citep{mohamed20}, which would further exacerbate current power imbalances in global health. On the other hand, increasing internet access and the democratization of machine learning knowledge and tools present an opportunity to create sustainable change in global health by empowering research partnerships to develop ML tools for locally relevant applications in health.

\paragraph{Contextualizing fairness:}To date, fairness in health research has furthered understanding and mitigation of biases across the machine learning development pipeline \citep{chenIY21, char18, siala22}. However most are contextualized to HICs. There are a few studies that have explored contextualizing fairness beyond the West.
 
Sambasivan et al. \citep{sambasivan21} utilizes mixed methods studies to identify subgroups, proxies and harms for fair ML in the context of India. For natural language processing applications, Bhatt et al. \citep{bhatt22} identify India-specific axes of disparities such as region and caste, and redefine global axes in an Indian context, such as gender, religion, sexual identity and orientation, and ability. Fletcher et al. \citep{fletcher21} provide detailed recommendations for fairness, bias and appropriate ML use in global health with examples using pulmonary disease classification in India.

\paragraph{Philosophical foundations:} Recent critical approaches to algorithmic fairness, e.g., by \citep{sambasivan21} and \citep{Kasirzadeh2022}, have problematized the philosophical scope of algorithmic fairness research, which thus far has largely been focused on distributive justice, i.e., justice which is concerned with the apportionment of benefits and burdens across members of a society. Such focus, it has been argued, is to the detriment of other important and relevant moral frameworks, including theories of social justice and non-western conceptions of normativity. The result is an inadequate approach to conceptualizing and implementing algorithmic fairness practices. We share this concern. Our contribution in this paper to globalizing fairness practices can be understood as part of an attempt to motivate the development of a more philosophically inclusive and contextually attuned approach to algorithmic fairness. In certain contexts, such as in certain non-western settings, fairness may require that ML-based interventions take into account factors which matter beyond the narrow question of how to fit such interventions into a fair distributive scheme. As the qualitative research results of our paper suggest, one such factor may be the history and legacy of colonialism.

\paragraph{Contributions:} This paper builds on previous work \citep{sambasivan21, bhatt22, fletcher21} by exploring fairness attributes for global health, with Africa as a case study. Throughout the paper, we will use ``fairness attribute'' to mean historical, demographic, or social properties (e.g., religion, age, socio-economic status, etc.) which may constitute ``axes of disparities'' in that they may serve as focal points around which disparities in the performance of ML-based interventions between Africa and the West may be measured. Axes of disparities are factors that run across the entire machine learning pipeline including problem selection and deployment considerations, whereas fairness attributes are a subset of the axes of disparities more specific to the data and model development. We make the following contributions:
\begin{itemize}
    \item We propose  (1) axes of disparities between Africa and the West, and (2) global axes of disparities in the context of Africa, identifying fairness attributes that should be considered with respect to Africa,  and delineating real world ML for health applications where these should be considered (section \ref{sec:axes_of_disparities}).
    \item We conduct a qualitative study on the relevance of these attributes and on the impact of colonialism on the development and deployment of ML systems in health in Africa (section \ref{sec:qualitative_study})
    \item Finally, we discuss contextual challenges to the application of ML in health in Africa and provide recommendations for fairness-aware ML solutions for health in Africa (section \ref{sec:challenges}).  
\end{itemize}

While there are common themes that may resonate across the continent, \textbf{Africa is a diverse continent} and there are attributes and axes of disparities which may apply to some countries and not to others.

\section{A Scoping Review of Fairness Attributes in Consideration for Africa}
\label{sec:axes_of_disparities}
 In this section, through a scoping literature review, we identify axes of disparities on a global scale, a continental scale, a local scale, and an individual scale. Common axes across most countries include colonial history (90\% of sub-Saharan Africa \citep{appiah10}), race (80\% fall under the definition of racially black \citep{wpr10}), and national-income levels (85\% of countries fall under world bank categories of low income, or lower-middle income \citep{worldbank24}). Within the continent there are inter-country disparities, for example national income levels (e.g., 6 countries are upper-middle income and only Seychelles is high income \citep{worldbank24}), the North African versus sub-Saharan African divide \citep{bentahar11}. There are also intra-country disparities across individual income levels, the rural-urban divide, ethnicity, and ethnic language, which we discuss in detail. We intentionally do not look at one specific “homogeneous” region on the continent in order to highlight unfairness that can affect the continent as a whole, as well as different levels of unfairness at the sub-regional and national levels. These axes of disparities are defined with the African context in mind, but the underlying logic can be applied to different geographical regions. We define key terms in Appendix \ref{A1}.
 \begin{figure*}[htbp]
  {\includegraphics[width=.8\linewidth]{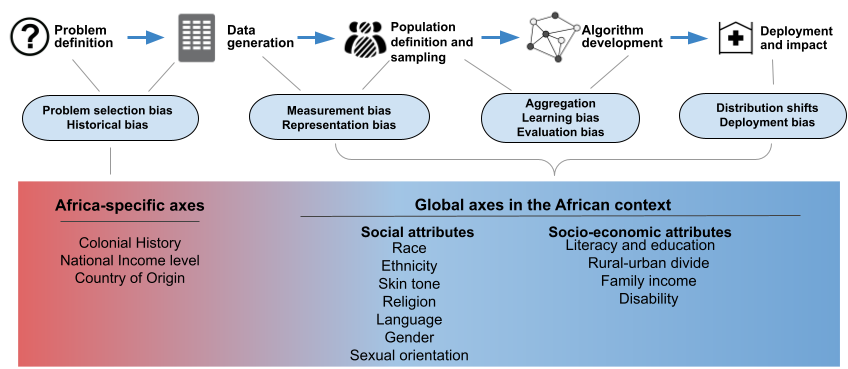}}
  \label{fig:pipeline}
  \caption{Categorization of African-contextualized axes of disparities with the type of biases they can induce along the machine learning pipeline. ML pipeline and biases figure modified from \citep{chenIY21} and \citep{suresh21}.}
  \vspace{0.1 cm}
\end{figure*}

\begin{table*}
  \label{tab:layman_survey_questions}
  \caption{Summary of the proposed axes of disparity. They are not mutually exclusive and some are connected}
  \begin{tabular}{l|lll}
    \toprule
    Country-level axes of disparities & \multicolumn{3}{c}{Individual-level axes of disparities} \\
    \midrule
     Colonial history\textsuperscript{†}& Race\textsuperscript{†,§} & Rural-urban location\textsuperscript{‡} & Religion\textsuperscript{†,§} \\
     National income level (NIL)\textsuperscript{†} & Ethnicity\textsuperscript{†,§} & Literacy and Education level\textsuperscript{‡,§} & Age\textsuperscript{‡,§} \\
     Country (origin\textsuperscript{‡} and residence\textsuperscript{‡})  & Language (ethnic)\textsuperscript{‡,§} & Skin tone\textsuperscript{†,§} &  Gender\textsuperscript{‡,§}\\
     Language (official)\textsuperscript{‡} & Sexual orientation\textsuperscript{†,§}   & Socio-economic status\textsuperscript{‡,§} & Disability status\textsuperscript{‡,§} \\
    \bottomrule
  \end{tabular}
  \begin{tablenotes}
    \setlength\labelsep{0pt}
    \footnotesize
            \item[1]\textbf{Legend:} ‡:Historically collected and/or used, †: accessible but not historically collected and/or used, §: Sensitive.\\  
  \end{tablenotes}
\end{table*}

\subsection{Axes of disparities between Africa and Western Countries}
\noindent\textbf{Colonial history:} There is strong evidence that Africa's colonial history and resulting structural challenges continue to create power imbalances between the formerly colonized on one hand and former colonial rulers and beneficiaries on the other \citep{eichbaum21}. 
Colonial history has been put forward by several scholars as a social determinant of health \citep{turshen77, ramos22,czyzewski11} and this has implications for data collection. We propose colonial history as a fairness attribute that could cause historical, representation, measurement, learning, evaluation, and deployment biases, and one that should be considered throughout the ML pipeline i.e. problem formulation, data collection, model development, and deployment. Learning bias is especially of note as objective measures for model performance should not be limited to accuracy, but should also seek to improve impact metrics. 
 
\noindent\textbf{National income level (NIL):} Ranked by NIL, Africa comprises  24 Low-income countries (LIC), 17 lower-middle income countries (LMIC), 6 upper-middle income countries, and 1 high-income country \citep{nada22}. 
Socio-economic disparities resulting from the disproportionately high number of LICs and LMICs on the continent imply limited funding from governments for research in ML for health, and limited availability of clinical data-generating devices and resources for ML development \citep{nabyonga21}. 
This results in problem definitions that are more in line with external funding requirements than with local needs\citep{nabyonga21}. Infrastructure limitations also impact access to data generation medical devices (e.g., mammograms), as well as electronic health storage systems, which can lead to representation bias and limited technical (e.g., compute) and personnel resources for ML development.

\noindent\textbf{Country of origin:}
As discussed above, different countries have different developmental and national income levels, and varying methods of implementing health strategies \citep{nabyonga21}, which are influenced by each country's leadership and political landscape. Hence machine learning applications that work in one country may not necessarily transfer to the other. Aggregate bias that may result in a one-size fits all model should be avoided by understanding implications for fairness that may be country-specific. 

\subsection{African-contextualized global axes of disparities}
We provide a contextual framing of globally applicable axes of disparities and illustrate why they should be considered both in importing ML models from outside Africa as well as when developing ML models within the continent.

\noindent\textbf{Race:} Race in fairness typically refers to structural racism referenced as a social construct due to the history of slavery and subsequent racial disparities in countries like the United States. Such a construct does not apply directly to Africa. 
The majority of Africans (80\%) are racially black and are subjugated to global anti-black rhetorics \citep{pierre13}. North African Arabs are externally racially white,  as are Afrikaans in South Africa. There are also people of Asian origin, predominantly in the east and southern Africa.  
However, most Africans do not self-identify by race and are more likely to identify by nationality and ethnicity \citep{asante12, maghbouleh22}. Demographic health questionnaires do not typically ask for ``race," though persons of African descent may be at higher risk for certain diseases.
 
One may consider racial disparities that impact Africans on a global scale, or within different countries in Africa, between Black and White North Africans \citep{pierre13}, or between Black and White South Africans \citep{verwey12}. This has implications for socio-economic disparities in health access, representation bias, and for race-based disparities in ML model performance.

\noindent\textbf{Ethnicity:} Africa has over 3000 ethnic groups and is the most genetically diverse continent. Unlike in HICs, most Africans associate with an ethnic group. Ethnicity, in addition to physical traits, language, and culture, defines "historical experience, aspirations, and world view" \citep{deng97}. 
 
Socio-economic disparities in accessing health may run along ethnic lines \citep{franck12}. This can lead to ML applications skewed towards dominant ethnic groups resulting in ethnic inequities.

\noindent\textbf{Religion:}
Africans practiced traditional religions pre-colonialism. With the introduction of Western religion, Christianity and Islam have grown to become dominant religions on the continent. Religion can be associated with ethnicity and socio-economic status, leading to ``ethno-religious disparities" in health \citep{gyimah06, ha14}. Religion can also impact access, perceptions, and adoption of health practices \citep{white15, schmid08, obasohan14}.  
Religion becomes a sensitive attribute when harmful stereotypes and inequities tied with ethnicity are propagated in machine learning models.

\noindent\textbf{Language:}
There are over 2000 languages in Africa, and most Africans speak more than one \citep{heine2000}. Oral language in Africa typically has the format of a dominant colonial language such as English, French, Arabic, or Portuguese \citep{obondo07}. 

These dominant languages, particularly colonial languages, facilitate cross-ethnic interaction, are typically both written and oral, are utilized in education and professional life, and have "led to linguistic inequality" \citep{obondo07}. 
On a per-country level, disparities occur between countries with English as an official language and others due to the "political, economic, and educational...power of the English language" \citep{plonski13}.

\noindent\textbf{Skin tone:}
On a global scale, skin tones of racially black Africans, light or dark, are associated with blackness and instigate anti-blackness discrimination \citep{pierre13}.

Among black Africans, lighter skin tone may be perceived as enabling better social status and favors \citep{Nyoni21}. As a result, 40\% of people on the continent practice cosmetic skin bleaching \citep{asumah22}. 
Bleaching in itself is a public health concern, increasing risk for severe health problems \citep{lewis12}. Within algorithmic development, studies have found algorithmic biases against darker skin tones \citep{buolamwini18, dooley21}.  In applications to health, this has implications for biases towards lighter skin which may be better represented in training data and cause discrepancies in dermatology applications, such as skin disease detection \citep{daneshjou22}.

\noindent\textbf{Gender:}
Historically patriarchal African culture, reinforced by colonial legacies, has led to inequitable distribution of wealth between genders \citep{jaiyeola20}. Female household and child caring duties, as well as professional duties more recently, further reduce access to basic health amenities for African women \citep{anaman20}. 
Transgender and gender non-conforming persons in Africa are stigmatized and discriminated against, with few health programs catering for transgender persons \citep{luvuno17}. This also has implications for health access and treatment, especially for sexually transmitted diseases \citep{merwe20}.

\noindent\textbf{Sexual orientation:}
Thirty-two countries criminalize homosexuality and it is punishable by death in 3 countries \citep{lars22}. Homosexual persons are stigmatized and discriminated against which causes exclusion and marginalization within health systems \citep{sekoni22, ross21}. 

\noindent\textbf{Literacy and Education level:}
Literacy and education levels impact access to care, health seeking behaviors, and understanding of health information, especially when delivered digitally \citep{amoah18}. This has implications not only for one's own health, but also for the health of one's children \citep{byaro21}.

\noindent\textbf{Age:}
Age is a global fairness attribute used both in machine learning model development and evaluations for health, due to age-specific incidence rates and co-morbidity \citep{Mhasawade21}.  

\noindent\textbf{Rural-urban divide:}
People living in rural areas may have disproportionate levels of lower socio-economic status, literacy, education, and limited access to health facilities \citep{oloyede17, sven15}. This makes people in rural areas most vulnerable to unethical machine learning practices.

\noindent\textbf{Socio-economic status:} Individual socio-economic disparities underlie most of the above axes of disparities mentioned. Africa has the second highest wealth distribution gap \citep{seery19}. This runs the risk of machine learning models in health perpetuating unfairness in deployment access or inaccuracies predominantly towards poorer persons.

\noindent\textbf{Disability:}
10\% to 20\% of African populations are affected by disabilities. Disabilities can exacerbate most of the attributes listed due to stigma and inadequate resources and policies at the country level \citep{adugna20, mckinney21}.

\section{A Qualitative Study on the perceptions of the fairness attributes and the connections between colonialism, health, and AI in Africa}
\label{sec:qualitative_study}
We conduct a qualitative investigation with experts and general population participants to study perceptions of the proposed axes of disparities as potential sources of bias in the development and deployment of machine learning models. We focus on colonialism as a central attribute of interest. The study and findings serve to provide evidence for the pre-identified fairness attributes, and how they can be accounted for when building ML-based solutions in health in Africa. We specifically sought to answer the questions 1) \textit{Are disparities along axes of the proposed attributes likely to be sources of bias for ML models?} and 2) \textit{Does colonialism have an impact on the implementation of AI solutions for health in Africa?}
In the rest of this section, we describe the study design, participant selection, and findings. To accommodate the diversity of backgrounds in the participants, we used AI and ML interchangeably in the survey and interviews. 
\begin{table*}[tb!]
\caption{Demographic distribution of general population (n=672) and expert (n=57) survey participants \vspace{-,3cm}}
\begin{tabular}{p{0.55\linewidth} p{0.42\linewidth}}

  \begin{minipage}{\linewidth}
  \begin{tabulary}{\textwidth}{CCCCC|CC}
    \toprule
    & \multicolumn{4}{c}{Age} & \multicolumn{2}{c}{Gender} \\
    \midrule
      & 18-20 & 21-29 & 30-39 & 40+ & Male & Female \\
    \midrule
    Gen Pop & 8\% & 49\% & 30\% & 13\% & 60\% & 40\% \\
    Experts & 0\% & 10.5\% & 54.4\% & 35.1\% & 57.9\% & 40.3\%\\
  \bottomrule
  \end{tabulary}
    \end{minipage} &

  \begin{minipage}{\linewidth}
  \begin{tabulary}{\textwidth}{CCCCCC|CCCC}
    \toprule
    & \multicolumn{5}{c}{Country}& \multicolumn{4}{c}{Race}\\
    \midrule
    & SA & Ken & Nig & Gha & Rwa & Bla & Asi & Whi & Oth\\
    \midrule
    Gen Pop & 128 & 125 & 169 & 125 & 125 & 94\% & 1\% & 4\% & 1\%\\
  \bottomrule
  \end{tabulary}
    \end{minipage} 
\end{tabular}
\\
    \begin{tablenotes}
\setlength\labelsep{0pt}
\footnotesize
        \item[1]\textbf{Legend:} SA = South Africa, Ken = Kenya, Nig = Nigerias , Gha = Ghana, Rwa= Rwanda, Bla = Black, Asi = Asian, Whi = White, Oth = Other 
    \end{tablenotes}
    
\end{table*}

\subsection{Methods}
\subsubsection{\bf Expert survey and in-depth interviews (IDIs):}
We recruited experts in various related fields using online flier distributions and word-of-mouth. Eligibility criteria were 1) researchers in ML, public health experts, health equity activists, entrepreneurs in digital health in Africa, and policy makers working on or in Africa, 2) 5+ years of experience in the relevant fields of work, 3) ability to read and understand English, 4) no affiliation with the institution conducting the research. Interested participants filled out a screener form, and those who met our eligibility criteria were accepted for the study. Participants digitally signed an informed consent form prior to beginning the study. Fifty-seven experts filled out a survey on Qualtrics in which they answered questions related to their perceptions on the proposed fairness attributes. The survey included definitions of machine learning, AI, fairness and bias.

A subset, twenty-eight of the experts, completed an in-depth interview via online video call in which they were asked about the impact of colonialism on AI in Africa, unique challenges of colonized countries with respect to AI implementation, and social dangers and benefits specific to colonized countries. To analyze the interviews, we used an inductive, semantic approach: three researchers independently extracted relevant and interesting phrases, grouped them under codes, then discussed and identified general themes from combining the codes. The full list of expert survey and interview questions, additional representative quotes, and demographic distribution can be found in Appendix \ref{A3}.

\subsubsection{\bf General population survey}

We surveyed 672 general population participants across 5 countries in Africa (South Africa, Kenya, Nigeria, Ghana, and Rwanda). Surveys were administered by a 3rd party vendor, Gutcheck, and were conducted in English in November 2023. Eligibility criteria were 1) have attained at least a high school diploma or equivalent, 2) ability to read and understand complex and technical text in English “very well,” and 3) be at least somewhat familiar with the term “AI” or ``ML.'' Participants were recruited, consented and compensated by the vendor. The survey questions probed participants on the connections between AI and colonialism, and on whether the proposed axes of disparities would lead to disparate performance by AI tools. The survey included definitions of ML, AI, fairness and bias. The full list of questions and detailed participant statistics can be found in Appendix \ref{A2}.

\subsection{Results}
\begin{figure*}[htbp]
  {\includegraphics[width=1\linewidth]{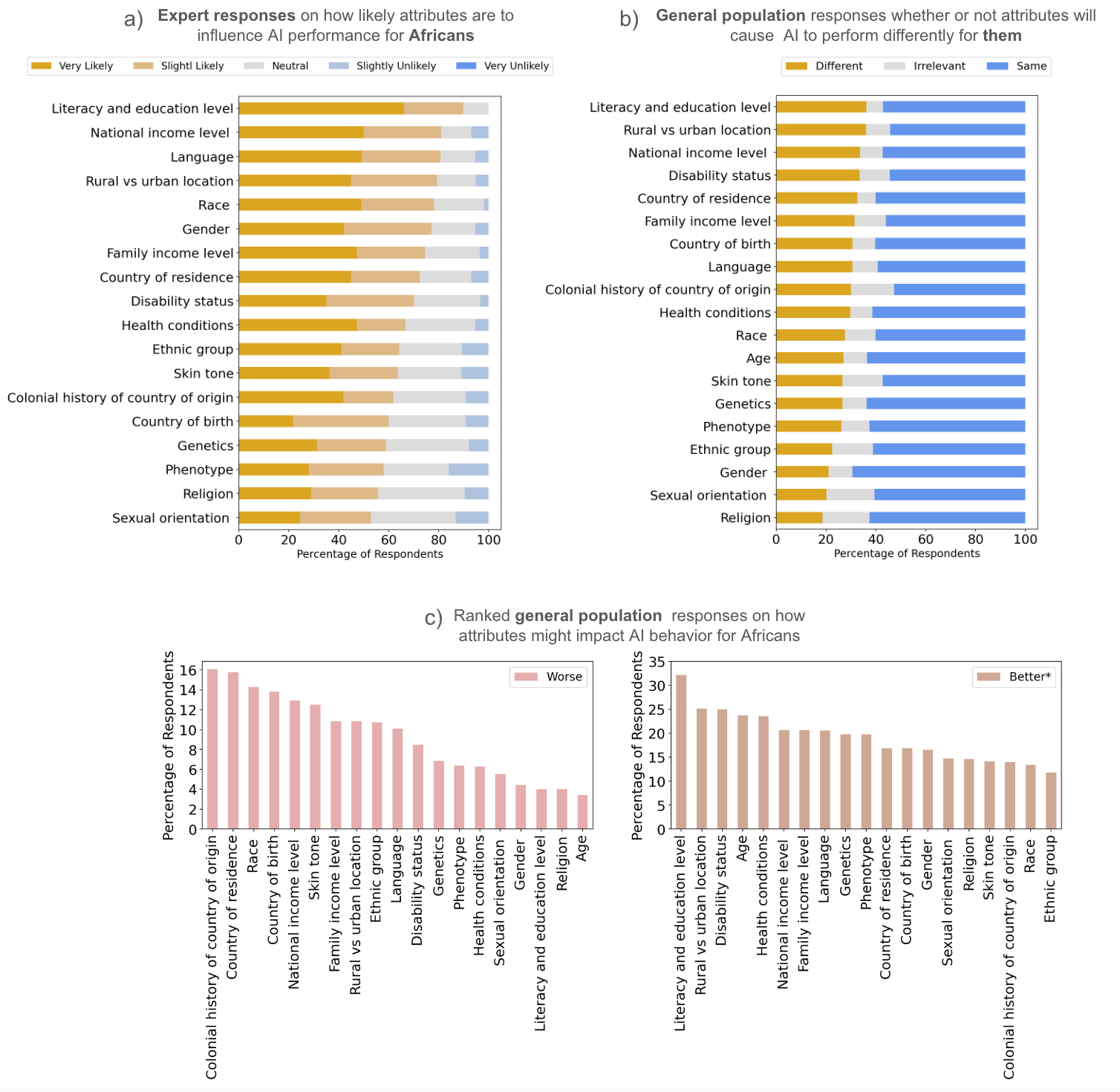}}
  {\caption{Perceptions of biases for machine learning tools and their associations with African-contextualized by axes of disparities. a) Experts' (n=57) responses on how likely attributes are to influence AI performance for Africans, b) General population responses (n=672) on whether attributes would cause AI to perform differently (better or worse) or the same for them as for others. They could also indicate whether the attribute was irrelevant. c) Breakdown and ranking of general population participant responses on which attributes would perform worse or better for them. *Better is not necessarily good, it reflects that our participant population is better off than other people in the same country. e.g., Most of our survey participants are skewed towards high literacy and education levels compared to the general population, and so AI may perform better for them than for others.}
  \label{fig:attributes}}
  \vspace{-.3cm}
\end{figure*}

\subsubsection{\bf Expert and general population opinions on disparities along the axes of the proposed attributes as sources of bias for AI-based tools in health in Africa:\\}

Most \textbf{expert respondents} indicated in a survey that the proposed axes of disparities would impact performance of an AI tool for Africans (Fig. ~\ref{fig:attributes} a). Each axis mostly ( 52\% - 90\%) received responses of being either very likely or extremely likely to cause machine learning models to perform differently for Africans, and none of the proposed axes received an unlikely response from experts. Literacy and education level, national income level, language, rural versus urban location, and race were the top five attributes experts identified as likely to impact AI performance and biases. As one expert put it, \textit{"It [AI] would perform differently for people who do not speak major languages...as their first languages or write it colloquially using their country's version...for men versus women...those who come from ethnic minority tribes particularly marginalised groups...darker skin tones and /or people with different skin from the majority in their context... different religious/cultural beliefs from those that align with predominant Western ideas...it would fail to discern between the subtleties across ethnic groups/tribes and countries and the algorithm would prioritise the experiences of countries with large populations like Nigeria and Ethiopia."} --[Public Health researcher, Female, Nigeria, 30-39]. 
See appendix \ref{A3} for additional 
representative quotes.\\

From the \textbf{general population} survey results (Fig. \ref{fig:attributes} b), we found that participants did not generally think AI tools would perform differently for them compared to others, along the proposed axes, which conflicted with expert responses. However, 18\% - 36\% of respondents indicated that axes of disparities would lead to disparate performance (either better or worse) for them. Only a small fraction indicated that the axes would be irrelevant. Literacy and education level, rural versus urban, national income level, disability status, and country of residence were the five top axes that participants indicated would cause an AI tool to perform differently for them. These were similar to expert results. Colonial history was the most likely to cause an AI tool to perform worse for participants. Literacy and education level were more likely to cause an AI tool to perform better. Given that our general population participants were skewed towards literate and educated backgrounds, this result was not surprising. However, this indicates that people who are not literate or educated could face adverse biases by AI tools. As one respondent indicates, \textit{"AI tools for health in Africa might perform differently across various demographic groups, including urban and rural populations, different age groups, varying socio-economic backgrounds, and distinct cultural communities. Factors like access to technology, healthcare literacy, and local health practices could also influence performance among these diverse groups"} --[Rwanda, 21-29, Woman]. Sexual orientation and religion were ranked lowest on likelihood for bias for both experts and general population participants. Broken down by country, general population respondents from South Africa were most likely to indicate that an AI tool would perform worse for them, which may be attributed to South Africa being more racially diverse and with black South Africans having most recently been subjugated to apartheid. See appendix \ref{A3} for additional quotes.

\subsubsection{\bf The impact of colonialism on the implementation of AI solutions for health in Africa:\\}
\begin{figure*}[htbp]
  {\includegraphics[width=.7\linewidth]{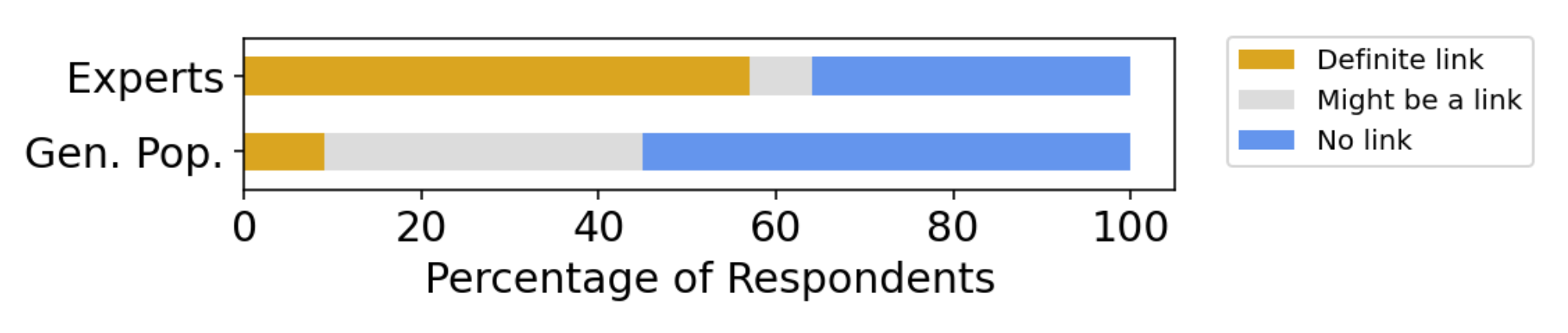}}
  {\caption{Responses by experts during IDIs ($n=28$) and general population participants surveys ($n=672$) on whether they see a connection between AI and colonialism. 57\% of experts found a definite link compared to only 9\% of general population participants.}
  \label{fig:gen_exp_bar}}
\end{figure*}

Opinions by both experts and general population respondents on the impact of colonialism on today’s applications of AI in healthcare are divided. Upon asking experts whether they believed that the history of colonialism impacts the application of AI in former colonies, 16 (57\%) said yes, 10 (36\%) said no, and 2 (7\%) indicated maybe (Fig. \ref{fig:gen_exp_bar}).
In the general population survey of 672 people across 5 countries, only 9\% thought there was a definite link, 36\% thought there might be a link and 55\% did not see a link between AI and colonialism (Fig. \ref{fig:gen_exp_bar}). South Africans were most likely to see a definite or slight connection (61\%) between AI and colonialism. There was a general consensus between experts on the importance of careful deployment of AI initiatives in order to increase trust and adoption. 

\paragraph{\bf Major Theme 1:}

A prominent theme that emerged was that colonial history has \textbf{``kept these countries behind’’} and \textbf{under-developed}, which can potentially lead to the \textbf{``lack of ownership’’} by the locals, \textbf{``economical dependency,''} and \textbf{``neocolonialists' exploitation of data and resources,’’} which in turn impact 
\textbf{trust}.

P1 puts it as \textit{``economic situation… ethnic strife, religious strife, … [are] byproducts of colonialism...[These were] tools for keeping people from uniting against colonizers … and because these countries have been kept behind...research in Africa is not digitized, it is not published in the high impact journals, it doesn't get to influence what is best practices of care...for AI. So our whole basis for science is biased by colonialism.”}
P29 discusses the \textit{``strong economical dependency...to former colonial powers which influences how technologies are adopted in former colonies. The imbalance of power can lead to AI being developed in only certain sectors that are of interest to former colonialists.'' }
Another expert discussing the historical self-interest agendas of the West says, \textit{"historically... the British medical system … wanted to treat the colonists [and made sure] British military officers were well educated against malaria and tropical diseases and [this education] wasn't meant for the Africans...But when they needed to employ Africans, it was cheaper and Africans were more resistant to African diseases obviously, so that's when it became an issue of interest to the West...so when there are international interventions, they[referring to Nigerians] worry...This was a position [based] solely on an exploitative economic venture by the colonialists‘’}-P6. Thus, some doubt the \textit{``altruism’’} that is now demonstrated in global health and AI for social good. P6 continues \textit{``[is it] solidly altruistic, free [data from] blacks, … [or] some \textit{economic games}, not for Africans?’’}. P12 adds \textit{“We have the Congo region…there's a lot of political instability…clearly influenced by the colonizers.”} Control of African leadership by colonial powers also comes into play, with P28 saying \textit{“we can't access some resources as fast as other countries, because of the way some of our leaders are still being controlled.”} This further influences mistrust \textit{“a few people might think that the West still wants to colonize us using AI. And so this will make an adoption or acceptance of AI products become a challenge”}-P20. \textit{“...those tools are ways in which our Colonial Masters still have a hold of us.”}-P10. 
There is also apprehension of neocolonialism through leadership - \textit{“There's this AI sector in our country, but it's mostly headed by British and French and it's completely biased towards the way they work and their culture”}-P5. 
\paragraph{\bf Major Theme 2:} Another predominant theme brought up by the participants indicating the impact of colonialism on AI Health application concerns \textbf{the existing biases that stem from imbalances in power amongst ethnic, religious and language subgroups due to colonialism and ties to colonial masters.}

For instance disparities in education and health centers by religion is brought up - \textit{“In countries like Malawi and Zambia for example, there is a really strong legacy and heritage of missionaries and their connection to hospitals...so some of the best hospitals were connected to churches”}. Ethnic disparities as a result of colonialist favoritism are also highlighted -\textit{“In Zambia...we have 72 tribes…maybe five or six of them which were sort of favored by the colonial powers….[As such the] western part of those countries are considered generally more educated than the rest of it of the country. So you find that taking technology or taking science of most kinds to the Western North Western southern part of the country is generally easy"} -P5 and P24 adds that \textit{“a lot of inequalities that exist between different ethnicities in general are a carry-on from colonialism, which is that there are more and more people who... based on their ethnicity actually receive different access to care and that is a consequence of colonialism to a large extent"}. The influence of colonial language also can result in disparities in education and technology adoption - \textit{“those who are colonized by the British tend to pick up English…being  able to speak English and those colonized by the French tend to have…official language being French, and that influences...research...provision of scholarships [with] preference given to...those colonized by the British”} -P12. P10 adds to this saying \textit{``technology in general is English-speaking driven…there's a chance that  Nigeria, Kenya would have more chances to move forward very fast [than] Senegal and Ivory Coast [which speak French]”}.  

\paragraph{\bf Other emerging themes on the potentially negative impacts of colonialism on AI:\\} 

There were also smaller but relevant themes around Western perceptions of Africa being ingrained into generative models - \textit{"most of the historical data used to build ChatGPT is the stuff that came from Google Books. It's like books as far back as two thousand years ago. Do I as a woman want to be judged culturally from books from 2000 years ago? No, I do not"}-P1, and around the continued use of frameworks inherited from colonial rule - \textit{"The legal framework the few countries still have some body of laws that are inherited from the colonial past and sometimes these laws have never been touched and are totally outdated."}-P10 
\textit{``But there are things that have ... created... some inequities into the system that we have not been able to fix ... after years of colonialism... There are ways in which we practice our medicine... that are based on how the set of doctors who are here, who are trained by whoever the colonial masters were and those behaviors... have seeped into how we still keep practicing medicine now... And so this is the way everybody passes it on to everyone. Now the issue with that is those inadvertently seep  into the way patients are treated. They seep into the data.''}-P22

\paragraph{\bf Major theme 3:}
There were a few experts who felt there was \textbf{no direct impact between AI and colonialism}, or that it was \textbf{complicated}.

Some participants talk about forgiveness and the potential of AI to avoid colonial biases if done correctly. \textit{“We do not bring out historical justices...Africans are more forgiving and welcoming...if the launch of AI within Africa is not done correctly, then they will try to associate it with colonial injustice. If  done correctly, I don't think Colonial history will be a hindrance.”}-P2.  One person indicated that colonialism happened a long time ago and there was a need to move on- \textit{“It's so long ago and we have advanced in our own ways better than before when we were colonized…we are free we can make our own decisions now.”} - P20. Whether or not the respondent’s country was colonized impacted responses as well. As P3 put it, \textit{“I am living in a country that has never been colonized before, I am from Ethiopia. So I don't think the colonial history of other African countries could have a negative impact on the implementation of AI”}.
A group of participants indicates the complexity of colonialism and its impact on AI applications, stating that this issue is very context-oriented and can vary across countries. \textit{“[within Nigeria] we don't have a homogeneous perception of how colonialism impacted Nigeria”}-P6. \textit{“I would say complex…situations are so different …maybe that the colonial term might be a bit too broad. We just say all African countries were colonized but…it's very different…the application of colonization…Outsiders came in and took resources... then the same outsiders came in and proposed solutions”} -P17. Others attributed limitations in AI not to colonialism but to leadership \textit{”the major problem is maybe leadership…corruption…that is depleting our resources”} -P29.

\paragraph{\bf Major Theme 4:} {\bf Looking forward–towards socially inclusive development of health applications of AI in former colonial spaces.}

Despite high associations of AI applications with the history of colonialism and neocolonialism in our expert IDIs, we saw some positive remarks towards making AI socially beneficial for Africa. \textit{“If AI can make Healthcare more accessible, then... it would be a social benefit.”} -P2. Another says \textit{``If for example, all the data in Nigeria is getting into one database…[for] viral infections, you can [attain] reduction in morbidity and mortality by allowing for public health measures''} -P7. P4 mentions colonized countries may benefit as \textit{"it will be easy to create an awareness on most of the colonial countries in Africa because they are already familiarized with the language in the Western countries such as English and French so that could be a positive impact to implement an AI in colonized African countries.”}

There were also forward looking strategies arising in interviews that suggested a careful adoption of AI applications. P24 specifically calls for responsible and accessible development of AI- \textit{"if AI is actually built responsibly…[considering] biases and instilled with the community that you're targeting in mind…to actually reach a lot of these communities…[make it] accessible”}. And others outline specific methods for preventing colonial tendencies, promoting agency and ownership of AI interventions by historically colonized people, and development of socially beneficial AI for health applications. \textit{“Issues to address include partnerships and networking...working with the local people...the learning institutions to build the trust so that it's not like a tool that you see when you don't work with the key stakeholders right from the start...if it is not so inclusive...[with] the ministries of health, because if they are continuously bypassed...ownership will not be there and then it will now be seen as those are things from the West...So building from scratch together with the communities builds trust, it builds the ownership of the development of that process...that will do away with the decolonizing of healthcare.” } -P4. However there were a few participants who  saw no path forward towards social benefits from AI for Africa without massive restructuring - \textit{“Nope, not without dismantling and deconstructing the systems and processes by which data is used to feed or build algorithms and predictive models. There are movements of AI for good but when you take a very critical lens...unpack those intentions, the very systems [are built] on structures of oppression and injustice. I think AI for social good would just be a myth”} - P26.

\section{Implications for machine learning development and practical recommendations}
\label{sec:challenges}
In this section, we expand on the themes brought up in the IDI interviews and characterize how they can manifest as unique challenges in the application of ML-based solutions in health in Africa. We end our discussion with practical recommendations for the development and deployment of fairness-aware ML-based solutions in health in Africa.

\begin{figure*}[htbp]
  {\includegraphics[width=.6\linewidth]{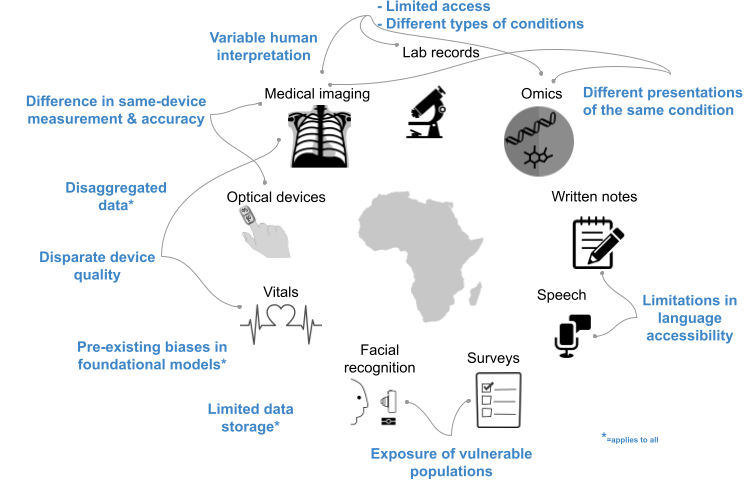}}
  \label{fig:modality}
    {\caption{African-contextualized barriers to ML for health by health modality. *=applies to all}}
\end{figure*}

\subsection{Structural challenges to the development of ML-based solutions for health in Africa}
In section \ref{sec:qualitative_study}, study participants highlighted the impact of colonialism such that many former colonies are "economically and technologically behind". This has a trickle-down effect on many health applications of ML, which we break down by health modality. Where applicable, we also connect these to the relevant axes of disparities and biases that may result.\\

\textbf{Electronic health record (EHR) systems:} Currently, machine learning requires digital data for model training or fine tuning. Machine learning for health in HICs have progressed as a result of large amounts of digital health data stored in patient EHRs. In most African countries, the majority of patient health records remain predominantly paper-based due to a lack of infrastructure and computational resources to keep digital copies \citep{Odekunleetal2017}. This phenomenon is exacerbated in countries with low NIL and in rural areas \cite{Akanbietal2012}. Awareness of the crucial role of having EHRs for ML has ushered in new proposals to digitize existing handwritten data and develop culturally appropriate solutions for EHRs. 

\textbf{Medical imaging:} Machine learning has shown the most promise for use in medical imaging tasks such as dermatology, pathology, mammography, ultrasound, CT, MRIs and X-rays \citep{castiglioni21}. However, imaging devices are limited in hospitals in Africa, due to their high cost and maintenance \cite{Geethanathetal2019, Ogboleetal2018}. Even when imaging devices are available, they are of varying quality and interpretation, which leads to poor generalization of ML models from HICs \citep{MANSON2023102653, Molluraetal2020, kawooya22}. They are also constrained to urban areas \cite{Ogboleetal2018, Anazodoetal2022}, which leads to disparities across the rural-urban divide \citep{kawooya22}, and are expensive which impacts socio-economic status. Additionally, medical images may not be digitally stored or connected to a patient's other health records due to EHR limitations discussed above. This can lead to limited data for retraining models and can generate representation and evaluation biases especially in instances when the data used to train the model is not representative of African populations, e.g., dermatology classifiers which do not work for darker skin tones.

\textbf{Lab values:}
sub-Saharan Africa has the lowest rate of high quality, accredited, clinical laboratories, and while there has been progress through various programs there are still gaps.  \citep{gershey10, alemnji14}. Access to laboratory testing services may be limited for people with lower financial means, and those in rural areas without labs who may have to travel to urban-based lab institutions \citep{petti06,castenda19}. Given limited EHRs, these results are not always integrated with a patient's comprehensive health profile, leading to segregated health information. In addition to data access limitations, limited infrastructure and personnel (e.g., pathologists) for lab test interpretation impact diagnostic efficiency and accuracy \citep{petti06, alemnji14}. These present unique challenges for ML purposes which are not observed in HICs. 

\textbf{Survey data:} Most application of machine learning to health remain tied to data collected in clinical settings. In the African context, however, there is a breadth of survey data, either self-reported, collected by community health workers, or in clinics as part of initiatives by different non-governmental organizations (NGOs) such as the USAID's demographic health surveys (DHS) \citep{dhs24}. These represent some of the largest, readily available longitudinal sources of health data from the continent and are used to provide data on disease trends over time by different regions and demographics. However surveys especially self-reported ones may be unreliable \citep{althubaiti16} and may also introduce a number of biases \citep{bradley21,althubaiti16}, which may impact people in rural areas, those of lower economic status, or those with lower literacy levels \citep{casale13}.  

\textbf{Unstructured written health notes:} Machine learning for unstructured health notes utilizes natural language processing models, and more recently large language models to extract information such as symptoms and action items, as well as provide disease classifications \citep{Irena20, li22}. These models are usually pre-trained on large amounts of text data from dominant languages and writing styles, which have been shown to exhibit biases. Using them without proper evaluation and fine-tuning may propagate these identified biases and lead to inaccuracies \citep{omiye23}. There also exist limitations to accessing written health notes in health facilities due to limited EHR availability.

\textbf{Medical speech:} Automatic speech recognition systems are used in various heath facilities by healthcare professionals to dictate notes without having to take time away from patient care \citep{brian22}. Large language models are further driving this application. Accents, style of speaking (e.g., pidgin), and literacy may impact speech recognition algorithms designed to be used in Africa \citep{dossou23, Tatman2017EffectsOT, koenecke21_spsc}. This can impact language disparities, and may vary by ethnicity, country of origin, literacy, and education. 

\textbf{Optical sensor devices:}
Optical sensor-based devices such as pulse oximeters and fitness trackers have been shown to have lower performance on darker skinned persons \citep{shi22}. Machine learning models developed for these devices may perpetuate measurement bias. This has direct implications for skin-tone disparities, which may be indirectly linked to country of origin and ethnicity.

\textbf{Omics:}
 Using ML to discover biomarkers from multi-omic data is a fast growing field, with potential for precision diagnosis, prognosis and prediction \citep{reel21}. Africa has large genetic diversity, but available omics data is underrepresented, making up only 1\% of omics databases \citep{hamdi21}. While the amount of data is increasing, facilitated by consortiums such as Human Heredity and Health in Africa (H3Africa) consortium and the H3Africa Bioinformatics Network (H3ABioNet), there remain limitations in labelled data, ethnic diversity representation, and integration with a patient's health history and profile \citep{hamdi21}. 

\subsection{Practical recommendations for fairness-aware ML solutions for health in Africa }

Addressing potential machine learning biases starts from structural and systemic changes in healthcare in Africa. (1) Access to basic health care for all persons, (2) increased government and private investment in health access, (3) health-based entrepreneurship and health-related research, (4) sustainable and context-specific development and deployment of health tools and technologies are essential for reducing disparities in health. Others more specific to machine learning and findings from our qualitative study are:

\textbf{Contextualization of the fairness criteria:} 
The nuanced differences in axes of disparities between HICs and LMICs lead to a fundamental need to define new or recontextualize existing fairness attributes within the local historical context. There is a unique opportunity to engage different stakeholders (e.g., researchers, policymakers, governance specialists, etc.) to define relevant fairness criteria in accordance with local laws and beliefs. An interdisciplinary approach at this stage is particularly important to avoid potential issues around competing definitions of fairness \citep{verma2018, Kleinberg16, Chouldechova17} and purely mathematical formulations that lead to unintended performance degradation for all groups \citep{levellingdown2023}.

\textbf{Collaborative problem selection:} As noted in \citep{chenIY21} and supported by our qualitative study results, biases stemming from unaligned incentives by external organizations, which lead to power imbalances, can severely affect priority areas, for example which diseases are studied. This has a trickle-down effect in terms of which data is collected regardless of relevance to the local population. It is particularly important at this step to consider the potential biases of stakeholders and ensure that voices of affected communities are included. 

\textbf{Awareness of historical biases towards foreign technology}: As our study finds, there is a general mistrust of technologies brought upon by former colonizers (\ref{sec:qualitative_study}). Awareness of this existing bias is critical in developing adequate deployment strategies to improve trust and adoption. Without this contextual understanding, deployment of ML-based solutions for health in Africa runs the risk of following historical trends and not taking into account cultural perspectives, under-representation, and other biases. 

\textbf{Caution around using pretrained models:} Given the scarcity of readily available large training data sets from Africa, a common ML approach consists of finetuning pretrained models. However, a prevalent concern lies within distribution shifts that may not be explicit, and the inability to adapt fairness properties to distribution shifts which are likely to occur \citep{schrouff2022diagnosing} when deploying models trained on Western-centric data to Africa. In dermatology diagnosis for example, presentations of the same disease are likely to differ between light skin and dark skin tones. Thus models trained on lighter skin may not work when deployed off-the-shelf. Ensuring generalization and transferability of fairness properties under distribution shifts is a new and active area of research \citep{schrouff2022diagnosing, giguere2022fairness, singh21, baldini-etal-2022-fairness, sadeghi20}. Use-cases specific to the African context can serve as motivating examples and drive impactful advances in this area.

\section{Discussion}

Initial algorithmic fairness metrics \citep{barocas-hardt-narayanan} were developed around US legal frameworks of disparate impact and disparate treatment \citep{barocas2016}. This centered the work on fairness around contextually relevant protected attributes such as race and gender \citep{Hutchinson2018}. We extend these criteria and reframe them in the African context for machine learning applications. Our human-subjects qualitative study amongst general population participants from 5 African countries, and experts across 17 different countries demonstrates this need and supports the axes of disparities we define. This is in line with previous work on fairness in India that demonstrated the need for recontextualizing fairness criteria \citep{sambasivan21, bhatt22, fletcher21}. We connect these axes to how they may impact AI for different health modalities and end by providing practical recommendations.

We conduct a deep dive into colonialism as an axe of disparity and while there have been a number of papers on colonialism and AI \citep{abeba20, mohamed20, racheladams21}, none of the previous work have conducted qualitative studies to gather expert and African general population opinion. This work demonstrates that experiences may impact Africans' perceptions/opinions about colonialism. People from countries and regions which were not colonized may not associate current challenges faced by health applications of AI with the history of colonialism. Among participants from formerly colonized spaces, there was also a divergence among perspectives with most experts in the IDI indicating a link between AI and colonialism. However in our survey, most of the general population indicated that there was not a definite link. Recommendations for moving forwards were raised by many of the participants, particularly around accessibility of AI tools, ownership, inclusion, and partnership with local people when it comes to their data. Given the potential connection between AI and colonialism, there is also need for a philosophically broader and contextually attuned approach to algorithmic fairness practices. Such an approach may take into account multiple frameworks for understanding the requirements of fairness–e.g., theories of distributive justice, of social justice, etc. Thus, in the case of Africa, if the history and legacy of colonialism impacts the application of AI in former colonies, then policy development specialists must work out the implications of this for ensuring the fair development, implementation, and distribution of ML-based interventions.

This work has implications for machine learning models, especially large language and multi-modal models which may perpetuate historical biases and may not have been trained on sufficiently representative data from Africa.\\

\textit{Limitations and future work:} There are a number of limitations in this study: (1) Because of the technicality of the survey questions and the need to be able to read and understand English, we exclude a significant amount of the population, especially those who would be most marginalized (low-literacy, rural location). Future work will address this by ensuring better representation, conducting surveys in different languages, and using speech-based methods. (2) We only conducted IDIs with experts and we will plan to also conduct IDIs with a subset of the general population for more nuanced responses. For future work, we also will explore how to operationalize the proposed attributes in existing bias mitigation techniques.

\section{Conclusion}
This work lays a foundation for contextualizing fairness beyond western contexts using machine learning applications to health in Africa as a case study. We perform a scoping review to identify axes of disparities in consideration for fairness in Africa and conduct a qualitative study with 672 general population participants and 28 experts participants across several African countries to validate their relevance. Our study also finds that there is a general mistrust of technologies imported from former colonizers among experts, and most associated resource constraints due to pre-existing economic and infrastructure inequities to colonialism. However, the majority of the general population participants surveyed did not think there was a direct link between AI and colonialism. Insights from these findings serve to better understand structural challenges to developing and deploying ML-based solutions for health in Africa as well as provide practical recommendations. We put forth this study as a starting point towards building fair machine learning in global health.

\section{Ethical considerations statement}

Each participant was provided informed consent and signed a consent form. Participation was completely voluntary and participants could withdraw at anytime. General population participants were recruited and consented through a vendor and were provided an incentive by the vendor, however we did our best in collaboration with the vendor to ensure appropriate incentive range. In-depth interview participants who completed the study were provided an incentive of 100 USD for participating.

\section{Researcher positionality statement}

All of the study's authors possess qualitative and/or quantitative research backgrounds, and have expertise in data science, artificial intelligence, machine learning, responsible AI for health and technology policy. The two lead authors are of African origin, as are three of the additional authors. Three of the study's authors are of South Asian and/or Middle Eastern origin. Two of the study's authors are native to the United States. Eight of the authors work on machine learning fairness-related areas in a USA-based corporation. The authors' work is shaped by their lived experiences in Africa and other formerly colonized spaces.

\section{Adverse impact statement}
We do not anticipate any adverse impacts from this work.

\bibliographystyle{ACM-Reference-Format}
\bibliography{sample-base}
\clearpage
\appendix
\onecolumn

\renewcommand{\thesection}{A.\arabic{section}}
\renewcommand{\thefigure}{A.\arabic{figure}}
\renewcommand{\thetable}{A.\arabic{table}}
\renewcommand{\theequation}{A.\arabic{equation}}

\setcounter{section}{0}
\setcounter{figure}{0}
\setcounter{table}{0}
\setcounter{equation}{0}

\section{Definitions}
\label{A1}
We define terms that are used in this paper. While there is not a standardized definition of fairness, we mainly follow the taxonomy described in \citep{Mehrabi2021}. We also provide other terms of general informational value for further understanding of fairness.\\

\begin{table*}[htb]
  \label{tab:definitions}
  \begin{tabulary}{0.9\textwidth}{|p{3cm}|p{10cm}|}
    \toprule
     Africa-specifc axes of Disparities &
    Disparities that primarily impact Africa. \\
    \hline
    Global axes of disparities in the context of Africa & 
    Disparities that have global implications but contextualized for Africa.\\
  \bottomrule
\end{tabulary}
\end{table*}

\begin{table*}[htb]
  \label{tab:definitions}
  \begin{tabulary}{0.9\textwidth}{|p{3cm}|p{10cm}|}
    \toprule
    Discrimination &
    `Discrimination can be considered as a source for unfairness that is due to human prejudice and stereotyping based on the sensitive
    attributes, which may happen intentionally or unintentionally'.\\
    \hline
    \hline
    Bias & `Bias can be considered as a source for unfairness that is due to the data collection, sampling, and measurement.' \\
    \hline
    Historical Bias &
    `Historical bias arises even if data is perfectly measured and sampled, if the world as it is or was leads to a model that produces harmful outcomes. Such a system, even if it reflects the world accurately, can still inflict harm on a population. Considerations of historical bias often involve evaluating the representational harm (such as reinforcing a stereotype) to a particular group'.\\
    \hline
    Representation Bias &
    `Representation bias occurs when the development sample underrepresents some part of the population, and subsequently fails to generalize well for a subset of the use population'. \\
    \hline
    Measurement Bias &
    `Measurement bias occurs when choosing, collecting, or computing features and labels to use in a prediction problem. Typically, a feature or label is a proxy (a concrete measurement) chosen to approximate some construct (an idea or concept) that is not directly encoded or observable. Proxies become problematic when proxy's are an oversimplification, are measured differently across groups, or accuracy differs across groups'. \\
    \hline
    Aggregation Bias &
    `Aggregation bias arises when a one-size-fits-all model is used for data in which there are underlying groups or types of examples that should be considered differently.A particular data set might represent people or groups with different backgrounds, cultures, or norms, and a given variable can mean something quite different across them. Aggregation bias can lead to a model that is not optimal for any group, or a model that is fit to the dominant population'. \\
    \hline
    Learning Bias &
    `Learning bias arises when modeling choices amplify performance disparities across different examples in the data. Issues can arise when prioritizing one objective (e.g., overall accuracy) damages another (e.g., disparate impact)'. \\
    \hline
    Evaluation Bias &
    `Evaluation bias occurs when the benchmark data used for a particular task does not represent the use '.\\
    \hline
    Deployment Bias &
    `Deployment bias arises when there is a mismatch between the problem a model is intended to solve and the way in which it is actually used'. \\
  \bottomrule
\end{tabulary}
\end{table*}

\begin{table*}[h!]
  \label{tab:definitions}
  \begin{tabulary}{0.9\textwidth}{|p{3cm}|p{10cm}|}
    \toprule
    Fairness & 
    `In the context of decision-making, fairness is the absence of any prejudice or favoritism toward an individual or group based on their inherent or acquired characteristics'. \\
    Individual Fairness & \citep{dwork2012}:
    `Individual fairness is captured by the principle that any two individuals who are similar with respect to a particular task should be classified similarly'.\\
    \hline
    Counterfactual Fairness & \citep{kusner2017}
    `The counterfactual fairness definition is based on the “intuition that a decision is fair towards an individual if it is the same in both the actual world and a counterfactual world where the individual belonged to a different demographic group.'\\
    \hline
    Group Fairness & 
    Broadly, group fairness notions aim to `treat different groups equally'. Below are two main statistical group fairness definitions that exist in the literature. \\
    \hline
     Equal Opportunity & \citep{hardt2016}:
     `This fairness notion requires that the probability of a person in a positive class being assigned to a positive outcome to be equal for both protected and unprotected (female and male) group members. In other words, the equal opportunity definition states that the protected and unprotected groups should have equal true positive rates'.\\
     \hline
     Demographic Parity & \citep{dwork2012}:
    ` requires that the overall proportion of individuals in a protected group predicted as positive (or negative) to be the same as that of the overall population'. \\
    \hline
    \hline
    Fairness in Relational Domains &
    `A notion of fairness that is able to capture the relational structure in a domain—not only by taking attributes of individuals into consideration but by taking into account the social, organizational, and other connections between individuals'. \\
    \hline
    Subgroup Fairness &
    Subgroup fairness intends to obtain the best properties of the group and individual notions of fairness. It picks a group fairness constraint like equalizing false positive and asks whether this constraint holds over a large collection of subgroups. \\
  \bottomrule
\end{tabulary}
\end{table*}
\clearpage

\section{Survey questions}
\label{A2}
\begin{table*}[h!]
  \label{tab:layman_survey_questions}
  \begin{tabulary}{0.9\textwidth}{L}
    \toprule
    General population survey questions  \\
    \midrule
    Q1. Do you see any connection between AI and colonialism? \\
    Q2. Do you believe that an automated AI tool would be biased towards or against you? In other words, do you have any reason to believe that an AI-based tool would treat you better or worse than other people? \\
    Q3. AI models may perform differently across different groups of people for a variety of reasons. How do you believe an AI tool for health would perform for you and based on the following criteria? \\
    Q4. Still thinking about how AI models may perform differently across different groups of people or different locations, what types of groups of people do you believe an AI tool for health in Africa would perform differently for? Please think any group of people, not necessarily a group you may consider yourself a part of. You may refer to the criteria above.? \\
  \bottomrule
\end{tabulary}
\end{table*}

\begin{table*}[h!]
  \label{tab:layman_survey_questions}
  \begin{tabulary}{0.9\textwidth}{L}
    \toprule
    Experts Survey questions\\
    \midrule
    Q1. Can you please categorize the following criteria by how likely they are to cause an AI tool for health to perform differently (be biased) for groups or sub-groups of people in Africa? \\
    Q2. Still thinking about how AI models may perform differently across different groups of people or different locations, what types of groups of people do you believe an AI tool for health in Africa would perform differently for? Please think any group of people, not necessarily a group you may consider yourself a part of. You may refer to the criteria above.\\
  \bottomrule
\end{tabulary}
\end{table*}

\subsection{Expert Participant distributions by country and specialty}
\begin{figure}
    \centering
    \includegraphics[width=\textwidth]{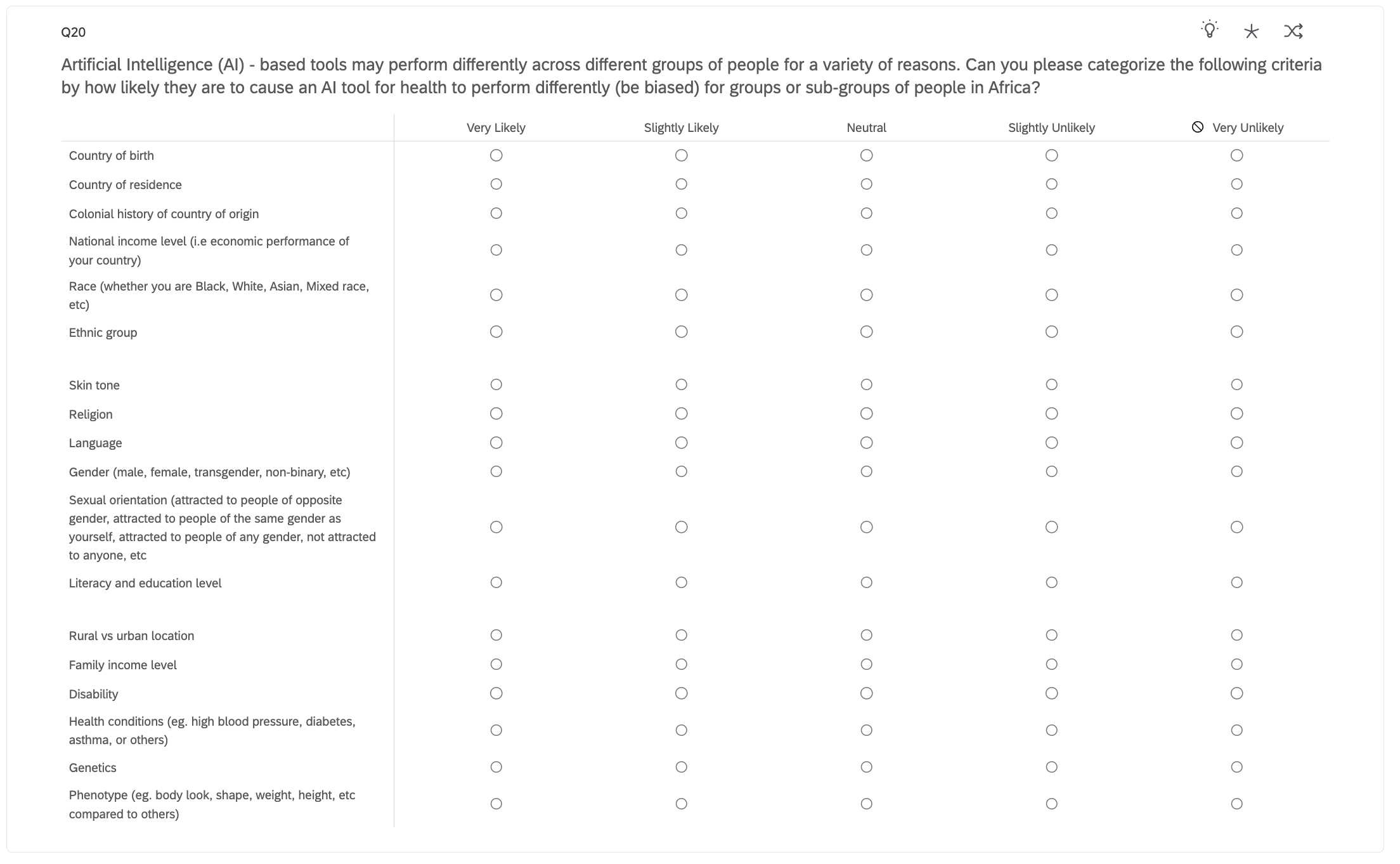}
    \caption{Example of the questionnaire in the survey}
\end{figure} 

\begin{table*}[h!]
  \label{tab:layman_survey_questions}
  \begin{tabulary}{0.9\textwidth}{L}
    \toprule
    Experts In-depth interview questions\\
    \midrule
    Q1. With respect to colonialism, is the application of AI to the health domain in African countries and former colonial territories impacted by the history of colonialism?\\ 
    Q2. Continuing to think about the impact of colonialism, what do you perceive as unique challenges in the application of AI to the health domain faced by African countries and former colonial territories? Probe: Are there specific challenges you can think of related to colonial history?\\
    Q3. With respect to colonialism, do you perceive any social benefits involved in the application of AI to African countries and former colonial spaces? If yes, what are some of the social benefits you can think of?\\
    Q4. With respect to colonialism, do you perceive any social dangers involved in the application of AI to African countries and former colonial spaces? If yes, what are some of the social dangers you can think of?\\
  \bottomrule
\end{tabulary}
\end{table*}
\clearpage

\section{Representative quotes from surveys on attributes likely to lead to bias}
\label{A3}
\begin{table*}[h!]
  \label{tab:layman_survey_questions}
  \begin{tabulary}{0.9\textwidth}{L}
    \toprule
    Representative quotes from general population:\\
    \midrule
    {\it "It may perform differently for Africans generally, but even worse for Africans living in the remote areas whose lifestyle is not yet understood lest alone have AI offer correct outcomes or diagnosis.”} {\bf -Ghana, 30-39, Man} \vspace{.2cm}\\
    {\it “AI tools for health in Africa might perform differently across various demographic groups, including urban and rural populations, different age groups, varying socio-economic backgrounds, and distinct cultural communities. Factors like access to technology, healthcare literacy, and local health practices could also influence performance among these diverse groups.”} {\bf -Rwanda, 21-29, Woman} \vspace{.2cm}\\
    {\it In general, there are a number of groups of people who might be disproportionately impacted by an AI tool for health in Africa. For example, people with low literacy levels or those who speak a minority language might have difficulty using the tool, and people living in rural areas might not have access to the necessary infrastructure to use it. People with certain disabilities or medical conditions might also be disproportionately impacted. Additionally, there could be gender or ethnic disparities in access to or use of the tool.”} {\bf -Nigeria, 30-39, Man} \\
  \bottomrule
\end{tabulary}
\end{table*}
\begin{table*}[h!]
  \label{tab:layman_survey_questions}
  \begin{tabulary}{0.9\textwidth}{L}
    \toprule
    Representative quotes from Experts\\
    \midrule
    \textit{"I think it would perform differently for people who speak/write do not speak major languages- English, French, Portuguese, or Spanish as their first languages or write it colloquially using their countries version (so a localised version). I think it would perform differently for men versus women, those who come from ethnic minority tribes particularly marginalised groups. I think that it would work very differently for people with darker skin tones and /or people with different skin from the majority in their context...it would perform differently for people who have different religious/cultural beliefs from those that align with predominant western ideas...it would fail to discern between the subtleties across ethnic groups/tribes and countries and the algorithm would would prioritise the experiences of countries with large populations like Nigeria and Ethiopia."} {\bf -Public Health researcher, Female, Nigeria, 30-39} \vspace{.2cm} \\
    \textit{"The poor and vulnerable would benefit a great deal from AI in healthcare but it's performance might not be great now. African society is religious and the rise of AI is causing agitation and some feel it's an insult to their beliefs. Having AI in health care will meet resistance in religious communities and affect uptake."}  {\bf -Clinical Officer, Kenya, non-binary, 30-39}\vspace{.2cm}\\
    \textit{People with albinism, Pregnant women, People with physical disability, People with mental health disorders, Migrants or displaced people or refugees, People in prisons."} {\bf -Policy maker, male,Malawi, 40-49} \\
  \bottomrule
\end{tabulary}
\end{table*}
\clearpage

\subsection{Expert Participant distributions by country of origin, country of residence and specialty}
\begin{figure}[!h]
    \centering
        \includegraphics[width=0.7\textwidth]{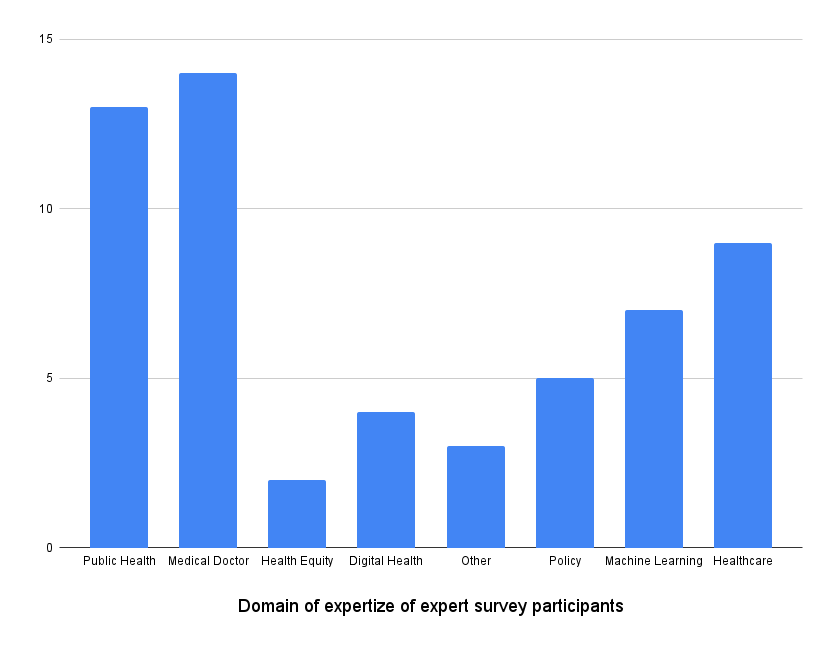}
        \caption{Expertize area of expert survey participants}
\end{figure} 

\begin{figure}
    \centering
        \includegraphics[width=0.7\textwidth]{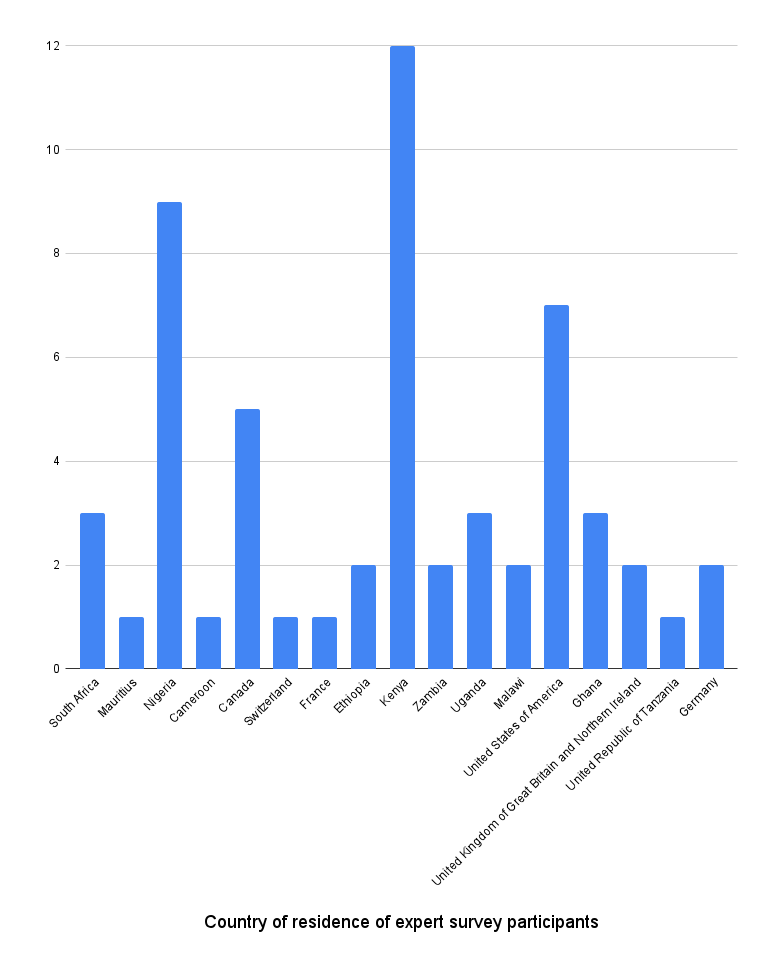}
        \caption{Country of residence of expert survey participants}
\end{figure} 

\begin{figure}
    \centering
        \includegraphics[width=0.7\textwidth]{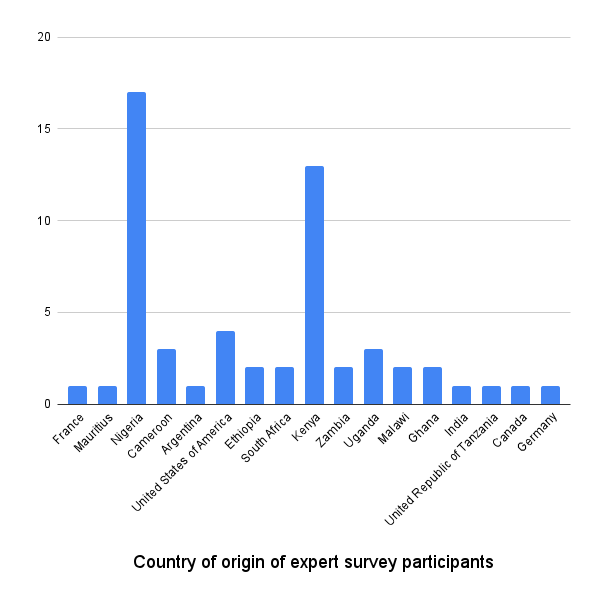}
        \caption{Country of origin of expert survey participants}
\end{figure} 
\clearpage

\subsection{Cover Picture showing a word cloud generated from responses from General Population survey when asked about axes of disparities }
\begin{figure}[!h]
    \centering
        \includegraphics[width=0.7\textwidth]{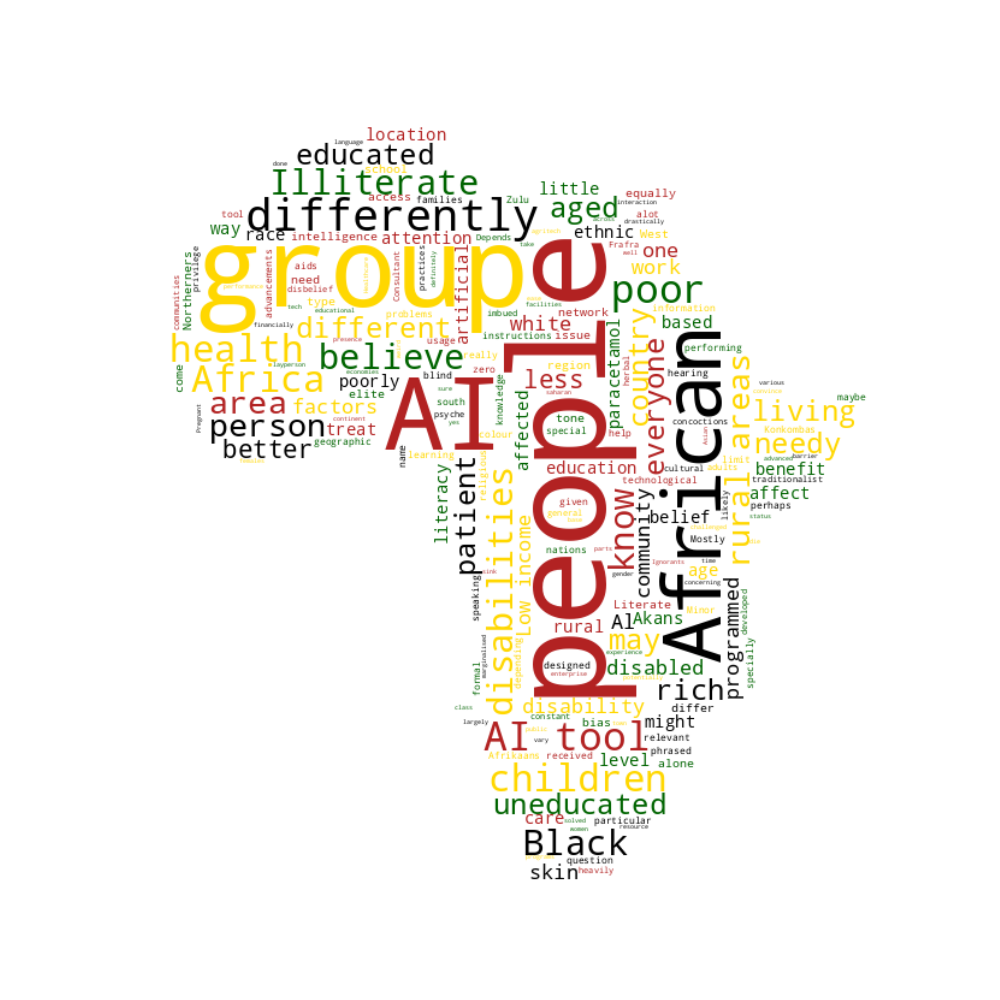}
        \caption{Word cloud of general population verbose responses to the question on axes of disparities}
\end{figure}

\end{document}